\documentclass{bmvc2k}

\title{SAMwave: Wavelet-Driven Feature Enrichment for Effective Adaptation of Segment Anything Model}

\addauthor{Saurabh Yadav$^*$}{saurabhy@iiitd.ac.in}{1}
\addauthor{Avi Gupta$^*$}{avig@iiitd.ac.in}{1}
\addauthor{Koteswar Rao Jerripothula}{kotesrj@iitk.ac.in}{2,1}

\addinstitution{
 IIIT Delhi\\
 New Delhi, India\\
}
\addinstitution{
 IIT Kanpur\\
 Kanpur, India\\
}

\runninghead{Yadav, Gupta, Jerripothula}{SAMwave}

\usepackage{times}
\usepackage{epsfig}
\usepackage{graphicx}
\usepackage{multirow} 
\usepackage{amsmath}
\usepackage{subcaption}
\usepackage{amssymb}
\usepackage{tikz}
\usepackage{nicematrix}
\usepackage{circledsteps}

\usetikzlibrary{shadings}
\usepackage{svg}
\usepackage{wrapfig,lipsum,booktabs}
\usepackage{xcolor,pifont}
\usepackage{colortbl}
\newcommand*\colourcheck[1]{%
  \expandafter\newcommand\csname #1check\endcsname{\textcolor{#1}{\ding{52}}}%
}
\newcommand*\colourcross[1]{%
  \expandafter\newcommand\csname #1check\endcsname{\textcolor{#1}{\ding{55}}}%
}

\newcommand{\cross}{\textcolor{red}{\ding{55}}}
\newcommand{\red}[1]{{\color{red}#1}}

\newcommand\crule[3][black]{\textcolor{#1}{\rule{#2}{#3}}}
\newcommand{\bluetext}[1]{\textcolor{blue}{#1}}
\newcommand{\redtext}[1]{\textcolor{red}{#1}}

\begin{document}

\maketitle
\def\thefootnote{*}\footnotetext{equal contribution}
\def\thefootnote{\arabic{footnote}}
\def\thefootnote{$\dagger$}
\footnotetext{Accepted to BMVC 2025. Please cite this work as \cite{samwave}}

\begin{abstract}
The emergence of large foundation models has propelled significant advances in various domains. The Segment Anything Model (SAM), a leading model for image segmentation, exemplifies these advances, outperforming traditional methods. However, such foundation models often suffer from performance degradation when applied to complex tasks for which they are not trained. 
Existing methods typically employ adapter-based fine-tuning strategies to adapt SAM for tasks and leverage high-frequency features extracted from the Fourier domain. However, Our analysis reveals that these approaches offer limited benefits due to constraints in their feature extraction techniques.
To overcome this, we propose \textbf{\textit{SAMwave}}, a novel and interpretable approach that utilizes the wavelet transform to extract richer, multi-scale high-frequency features from input data. Extending this, we introduce complex-valued adapters capable of capturing complex-valued spatial-frequency information via complex wavelet transforms. By adaptively integrating these wavelet coefficients, SAMwave enables SAM's encoder to capture information more relevant for dense prediction. Empirical evaluations on four challenging low-level vision tasks demonstrate that SAMwave significantly outperforms existing adaptation methods. This superior performance is consistent across both the SAM and SAM2 backbones and holds for both real and complex-valued adapter variants, highlighting the efficiency, flexibility, and interpretability of our proposed method for adapting segment anything models.

\end{abstract}
\section{Introduction}
\label{sec:intro}
Large vision models~\cite{DBLP:journals/tmlr/OquabDMVSKFHMEA24, clip, DBLP:conf/iclr/DosovitskiyB0WZ21} have become the dominant solution for numerous computer vision tasks, significantly advancing fields like semantic segmentation~\cite{DBLP:conf/iccv/KirillovMRMRGXW23, DBLP:conf/iclr/RaviGHHR0KRRGMP25} and object detection. However, their substantial parameter counts, while enabling high performance on their trained tasks, lead to sub-optimal generalization to other related problems, a challenge particularly evident in low-level vision tasks. For instance, while SAM~\cite{DBLP:conf/iccv/KirillovMRMRGXW23} excels at semantic segmentation, it struggles with tasks like camouflaged object detection. The difficulty of developing a single, universally high-performing model for all low-level tvision tasks has led to growing interest in adapting large models~\cite{DBLP:conf/iccv/RanftlBK21, DBLP:journals/pami/VandenhendeGGPD22, DBLP:journals/tmm/LuSDWL24, DBLP:conf/cvpr/LiLB22a, DBLP:conf/iccv/BruggemannKOGG21, DBLP:conf/nips/LiWZSJLDXT23}. Another common fine-tuning strategy is finetuning; however, it faces several challenges, including catastrophic forgetting~\cite{DBLP:conf/nips/ChenCLIK23, DBLP:conf/nips/HengS23, DBLP:conf/nips/ShinLKK17, Cao_2024_CVPR, DBLP:conf/nips/BabakniyaF0SA23, DBLP:conf/wacv/0005SDCWG24} and the reliance on large-scale datasets for adequate adaptation. 
\begin{wrapfigure}{l}{0.38\linewidth}
\centering
\includegraphics[width=.4\textwidth]{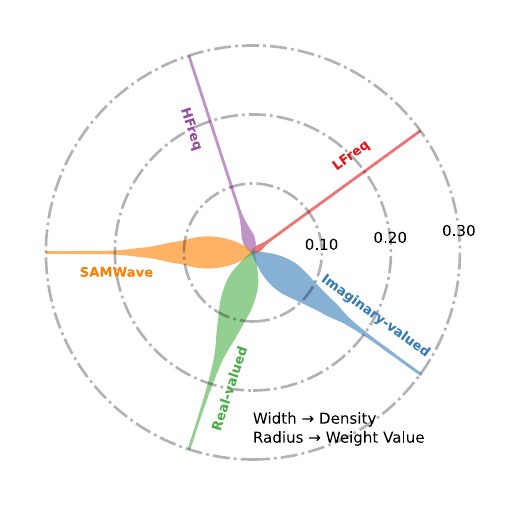}
\caption{\small Learned weights for previous high frequency extraction method, compared with SAMwave's real and complex adapter weights. Unlike previous method, SAMwave's weights are not concentrated around $0$.}
\label{fig:teaser}
\end{wrapfigure}\noindent In this work, we investigate the optimal adaptation of large models to related tasks, focusing particularly on \textbf{\textit{SAM}}, which comprises of a powerful encoder and a prompt-conditioned decoder. We hypothesize that the encoder can produce a sufficiently rich latent representation, such that  the decoder can generate effective segmentation mask for new task--given appropriate supervision. Previous works~\cite{DBLP:conf/cvpr/LiuSPC23, chen2023sam} have shown the efficacy of providing high-frequency features to the encoder for adapting a frozen decoder. These methods are typically performed by transforming input images into the Fourier domain and applying a predefined mask to emphasise high-frequency components. Different images possess varying frequency distributions, and a fixed mask leads to suboptimal feature selection. Our empirical study, presented in Tab.~\ref{tab:analysis}, supports this observation--demonstrating that simply inverting the mask yields comparable performance across tasks. Fig.~\ref{fig:teaser} further illustrates this issue. This motivates our departure from rigid, predefined Fourier-domain masks toward a more adaptive feature selection mechanism. To this end, we propose a novel high-frequency feature extraction method using wavelet transform, which offers superior spatial and frequency localization~\cite{DBLP:books/daglib/0001347, DBLP:conf/cvpr/RamamonjisoaFWL21, DBLP:journals/corr/abs-2402-19215}.  Specifically, we extract horizontal ($I_{lh}$), vertical ($I_{hl}$), and diagonal ($I_{hh}$) features from an image $I$, combining them into a high-frequency image $I_{HF}=I_{lh}+I_{hl}+I_{hh}$, which provides finer details and texture to the encoder.  Wavelet transform is advantageous due to its ability to capture both frequency and spatial information, allowing for a more precise representation of image details compared to Fourier-based methods. We further extend this approach with complex wavelet transforms and complex-valued adapters. Complex-valued representations offer a more comprehensive way to capture image information, potentially leading to improved feature extraction and adaptation, as demonstrated in various applications\cite{DBLP:conf/iclr/TrabelsiBZSSSMR18, DBLP:conf/iccv/YadavJ23}. We are the first to show the adaptation of vision foundational models using a complex-valued approach. Our contributions are summarized as follows: \textit{{\small \Circled[fill color=black, inner color=white]{\textbf{1}}} We demonstrate the limitations of using the Fourier domain with predefined masks for high-frequency feature extraction. {\small \Circled[fill color=black, inner color=white]{\textbf{2}}} We introduce a Wavelet High-Frequency (WHF) module that leverages the wavelet transform for effective fine-tuning. {\small \Circled[fill color=black, inner color=white]{\textbf{3}}} We propose a complex-valued extension of our method using complex wavelet transform and complex-valued adapters. {\small \Circled[fill color=black, inner color=white]{\textbf{4}}}  Our experiments show significant performance improvements over existing tuning methods for SAM and SAM2 across four low-level tasks.}

\section{Related Work}
\label{sec:related}
\paragraph{Visual Adapters.}The paradigm of adapting large pre-trained models for downstream tasks, initially popular in NLP \cite{DBLP:conf/icml/HoulsbyGJMLGAG19}, has effectively extended to computer vision \cite{DBLP:conf/eccv/LiMGH22, DBLP:conf/iclr/ChenDWHLDQ23}. Recent work has shown its applicability to vision foundation models like SAM \cite{DBLP:conf/iccv/KirillovMRMRGXW23}, notably through visual prompting methods \cite{DBLP:conf/cvpr/LiuSPC23, chen2023sam}. While prior adaptation methods often prioritize global context, our work focuses on incorporating crucial local, high-frequency information via wavelets to enhance SAM's performance on low-level vision tasks.

\paragraph{Low-level Vision Tasks.}Our work addresses several low-level vision tasks where fine-grained details are critical for accurate localization.
\textbf{Camouflaged Object Detection (COD)} involves distinguishing objects seamlessly blended into their background \cite{Fan_2020_CVPR, aixuan_cod_sod21, DBLP:conf/wacv/GuptaJT25}, requiring models to capture subtle boundary cues \cite{Pang_2022_CVPR, yin2022camoformer}.
\textbf{Defocus Blur Detection (DBD)} aims to identify out-of-focus image regions, relying on variations in blur kernels that are often captured by local features like edges \cite{KaraaliJ18, ShiXJ15, TangZLWZ19}.
\textbf{Shadow Detection (SD)} requires distinguishing shadows from true objects, often involving analyzing illumination patterns and local spatial context \cite{Hu0F0H18, Zhu0KL21}.
\textbf{PolyP Detection (PD)} involves delineating polyp regions from endoscopic images, thereby assisting clinicians in the early detection and treatment of colorectal abnormalities \cite{Trinh_2024_BMVC, DBLP:conf/cvpr/SharmaKJBDB22}.
\textbf{Forgery Detection (FD)} focuses on locating manipulated pixels, where inconsistencies in noise levels or texture often manifest as high-frequency anomalies \cite{DBLP:conf/eccv/CunP18, DBLP:journals/tifs/FridrichK12}. While some methods \cite{DBLP:conf/cvpr/LiuSPC23} use generic high-frequency components for low-level tasks, we propose a more principled approach using specific wavelet decompositions to provide targeted local feature guidance to SAM for these diverse low-level vision challenges.
\begin{figure*}
\centering
\includegraphics[width=\textwidth]{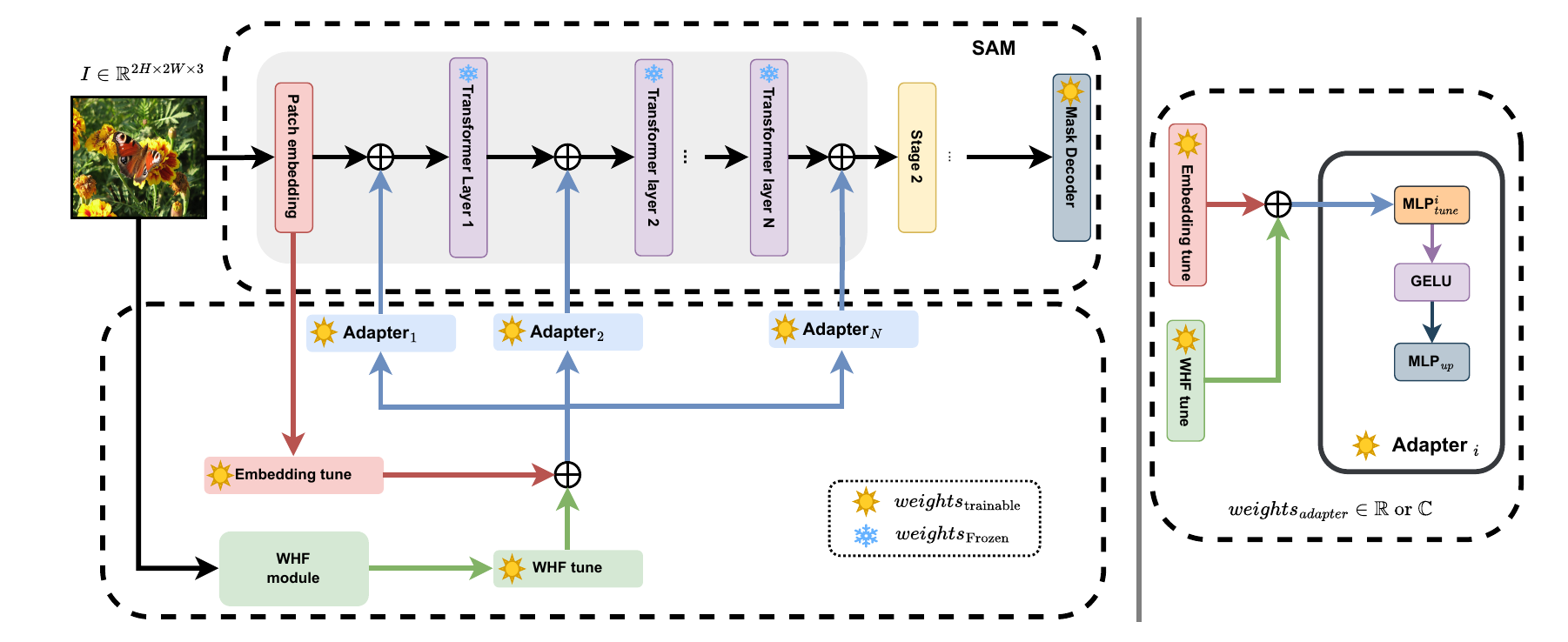}
\caption{\small Overview of SAMwave for low-level vision tasks. \textbf{Right} shows that the image is passed through SAM and our WHF (wavelet high frequency) module, where the high-frequency features are extracted. The extracted details are then combined with patch embeddings obtained from the transformer encoder. Note that the encoder is frozen; we only updated the decoder for our training. \textbf{Left} shows the overview of a single adapter used; for complex adapters, we replace the weights of MLP layers with complex values.}
\label{fig:architecture}
\end{figure*}
\section{Method}
In this section, we describe our proposed network SAMwave. The idea is to leverage the high-frequency features obtained from wavelet transform and use them to guide the Segment-Anything model (SAM) \cite{DBLP:conf/iccv/KirillovMRMRGXW23} for more precise segmentation in various tasks.

\newcommand{\greenbox}[1]{\colorbox{green!40}{#1}}
\newcommand{\bluebox}[1]{\colorbox{blue!30}{#1}}
\begin{table}[h]
\caption{\small Result of low frequency (LFreq) \& high frequency (HFreq) on camouflaged object detection on three separate datasets. We also show the difference in the performance of both approaches.}
\centering
\vspace{5pt}
\setlength{\tabcolsep}{3pt}
\resizebox{0.9\columnwidth}{!}{
\begin{NiceTabular}{l||cccc||cccc||cccc}
\toprule
\midrule
\multirow{2}{*}{} & \multicolumn{4}{c||}{CHAMELEON} & \multicolumn{4}{c||}{CAMO} & \multicolumn{4}{c}{COD10K} \\

                        & $\mathcal{S}_m \uparrow$    & $\mathcal{E}_\phi \uparrow$    & $\mathcal{F}^{w}_{\beta} \uparrow$    & $\mathcal{M} \downarrow$    & $\mathcal{S}_m \uparrow$   & $\mathcal{E}_\phi \uparrow$   & $\mathcal{F}^{w}_{\beta} \uparrow$   & $\mathcal{M} \downarrow$  & $\mathcal{S}_m \uparrow$    & $\mathcal{E}_\phi \uparrow$   & $\mathcal{F}^{w}_{\beta} \uparrow$   & $\mathcal{M} \downarrow$   \\ \midrule
     \Block[tikz={shade, shading=radial, inner color=lime!80, outer color=white}]{1-1}{}{HFreq} & \Block[tikz={shade, shading=radial, inner color=lime!80, outer color=white}]{1-1}{}{\textbf{0.896}} & \Block[tikz={shade, shading=radial, inner color=lime!80, outer color=white}]{1-1}{}{\textbf{0.919}} & 0.824 & \Block[tikz={shade, shading=radial, inner color=lime!80, outer color=white}]{1-1}{}{\textbf{0.033}} & \Block[tikz={shade, shading=radial, inner color=lime!80, outer color=white}]{1-1}{}{\textbf{0.847}} & \Block[tikz={shade, shading=radial, inner color=lime!80, outer color=white}]{1-1}{}{\textbf{0.873}} & \Block[tikz={shade, shading=radial, inner color=lime!80, outer color=white}]{1-1}{}{\textbf{0.765}} & \Block[tikz={shade, shading=radial, inner color=lime!80, outer color=white}]{1-1}{}{\textbf{0.070}} & \Block[tikz={shade, shading=radial, inner color=lime!80, outer color=white}]{1-1}{}{\textbf{0.883}} & \Block[tikz={shade, shading=radial, inner color=lime!80, outer color=white}]{1-1}{}{\textbf{0.918}} & 0.801 & \Block[tikz={shade, shading=radial, inner color=lime!80, outer color=white}]{1-1}{}{\textbf{0.025}} \\
     \Block[tikz={shade, shading=radial, inner color=orange!80, outer color=white}]{1-1}{}{LFreq} & 0.888 & 0.903 & \Block[tikz={shade, shading=radial, inner color=orange!80, outer color=white}]{1-1}{}{\textbf{0.830}} & 0.034 & 0.824 & 0.843 & \Block[tikz={shade, shading=radial, inner color=orange!80, outer color=white}]{1-1}{}{\textbf{0.765}} &0.078 & 0.872 & 0.901 & \Block[tikz={shade, shading=radial, inner color=orange!80, outer color=white}]{1-1}{}{\textbf{0.803}} & 0.027 \\ 
\midrule
     & \red{+ 0.008} &  \red{+ 0.016} &\red{- 0.006} &\red{+ 0.001} &\red{+ 0.023} &\red{+ 0.030} &\red{0.000} &\red{+ 0.008} &\red{+ 0.009} &\red{+ 0.017} &\red{- 0.002} &\red{+ 0.002} \\ 
 \midrule \bottomrule
\end{NiceTabular}
}
\label{tab:analysis}
\end{table}

\subsection{Motivation for High-Frequency Feature Utilization}
Precise segmentation, particularly the accurate delineation of object boundaries, is a challenging task that fundamentally relies on capturing fine-grained image details. These details are predominantly encoded within the high-frequency components of the image signal. Therefore, effectively leveraging high-frequency features is crucial for achieving high-quality segmentation masks.

Previous methods aiming to incorporate high-frequency information, such as those in \cite{chen2023sam, DBLP:conf/cvpr/LiuSPC23}, have often relied on extracting these features in the Fourier domain using a predefined, fixed mask. However, a significant limitation of this "one mask fits all" approach is its inability to adapt to the diverse characteristics and varying distributions of high-frequency information present across different images and tasks. Consequently, a static Fourier mask may not effectively isolate the relevant high-frequency cues necessary for robust fine-grained segmentation in all scenarios.

To investigate the efficacy of this fixed-mask approach for capturing essential high-frequency information, we conducted an empirical study. We compared the performance using features extracted via the standard high-frequency mask (HFreq) against those extracted by inverting this mask to obtain low-frequency features (LFreq). Our experiments focused on the challenging task of camouflaged object detection, and the results are summarized in Tab.~\ref{tab:analysis}. While HFreq shows superior performance on 9 out of 12 metrics compared to LFreq, a closer examination reveals that the average performance gain offered by HFreq is remarkably small, merely \textbf{0.009}. This marginal difference, despite high frequencies theoretically being more aligned with boundary information, suggests that a fixed Fourier mask is not an optimal strategy for extracting discriminative high-frequency features that significantly enhance segmentation performance across varied data. This observation motivates the need for a more adaptive and effective approach to harness high-frequency information for guiding models like SAM towards more precise segmentation.

\subsection{SAM as the Backbone}
\label{sec:sam}
Our SAMwave framework employs the Segment-Anything Model (SAM) \cite{DBLP:conf/iccv/KirillovMRMRGXW23} as its core backbone, leveraging its robust pre-trained capabilities for various segmentation tasks. SAM comprises three main components: an image encoder, a prompt encoder, and a mask decoder.

The image encoder, a pre-trained ViT-H/16, processes the input image $I \in \mathbb{R}^{H \times W \times 3}$ to produce dense embeddings $E_{\text{img}} \in \mathbb{R}^{\frac{H}{16} \times \frac{W}{16} \times 256}$. These embeddings are spatially downscaled by 16$\times$ and have 256 channels after a $1 \times 1$ convolution. The prompt encoder processes input prompts (such as dense features from our method) typically using convolutions, and these prompt embeddings are element-wise added to $E_{\text{img}}$. The mask decoder, a lightweight transformer-based network inspired by \cite{DBLP:conf/eccv/CarionMSUKZ20, cheng2021maskformer}, takes the combined embeddings and predicts the final segmentation mask via upsampling and a linear classifier.

In SAMWave, we freeze the weights of the pre-trained image encoder and train only the parameters of the mask decoder to adapt SAM's segmentation generation to the task-specific guidance.

\begin{wrapfigure}{r}{0.5\textwidth}
\centering
\includegraphics[width=0.55\textwidth]{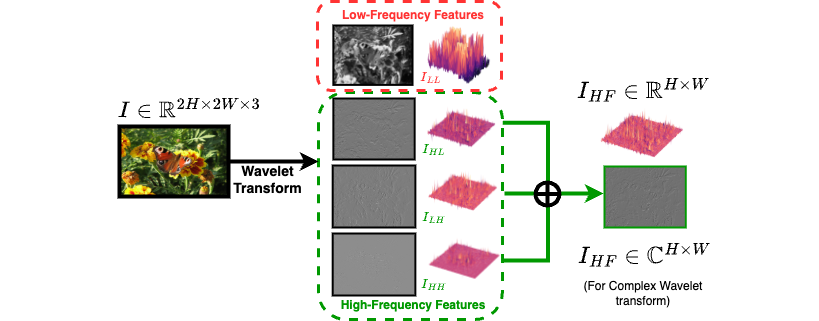}
\vspace{5pt}
\caption{\small An overview of the WHF module, we first use the wavelet transform to obtain high- and low-frequency features. Then, we combine the high-frequency features to create a feature map, which is then used to provide finer details in the image.}
\vspace{-7pt}
\label{fig:wavelet_arch}
\end{wrapfigure}

\subsection{Wavelet High-Frequency Module (WHF)}
\label{sec:wavelet}
The Wavelet High-Frequency (WHF) module extracts rich high-frequency spatial features using the Discrete Wavelet Transform (DWT). Given an input image $I \in \mathbb{R}^{H \times W \times 3}$, we first resize it to $I' \in \mathbb{R}^{2H \times 2W \times 3}$. This upsampling compensates for the DWT's inherent downsampling, ensuring output subbands match the original $H \times W$ resolution.

Applying a 2D DWT to $I'$ decomposes it into four subbands at $H \times W \times 3$ resolution: $I'_{ll}$ (approximation), $I'_{lh}$ (horizontal details), $I'_{hl}$ (vertical details), and $I'_{hh}$ (diagonal details).
$ I' \xrightarrow{\text{DWT}} \{I'_{ll}, I'_{lh}, I'_{hl}, I'_{hh}\} $
Focusing on fine details, we exclude the low-frequency $I'_{ll}$ subband and combine the high-frequency detail subbands ($I'_{lh}, I'_{hl}, I'_{hh}$) to form the composite high-frequency map $I_{HF}$.
$ I_{HF} = I'_{lh} + I'_{hl} + I'_{hh} $
$\in \mathbb{R}^{H \times W \times 3}$ captures edge and texture information (Fig.~\ref{fig:wavelet_arch}). This map is then projected to a $D_{embed}$ channel dimension.


A key advantage of this wavelet-based approach is its inherent flexibility. While we describe the process using standard real-valued DWT, the WHF module can seamlessly incorporate complex wavelet transforms, such as the Dual-Tree Complex Wavelet Transform (DT-CWT)\cite{Lou_2024_CVPR}. The core procedural steps---resizing, wavelet decomposition into subbands, selection and combination of high-frequency subbands, and linear projection---remain identical. The only change lies in the specific wavelet basis functions used in the decomposition, allowing for adaptation to properties captured by complex wavelets\cite{Saragadam_2023_CVPR} (\textit{e.g.}, phase information, shift invariance) without altering the module's overall structure.

\subsection{SAMwave Architecture}
\label{sec:samwave}
\textit{\textbf{SAMwave}} enhances SAM's segmentation capabilities by integrating learned high-frequency wavelet features \textit{via} lightweight adapters. As illustrated in Fig.~\ref{fig:architecture}, the architecture consists of three key modules--Embedding Tune, WHF Tune, and Adapters--which are integrated into the SAM backbone.

\noindent \textbf{Embedding Tune:} This module generates a learned spatial feature map $\psi_{PE} \in \mathbb{R}^{\frac{H}{16} \times \frac{W}{16} \times \gamma}$ from SAM's image encoder embeddings $\mathbf{F}_{\text{emb}}$ using a learnable MLP layer $\phi_{PE}(.;\theta_{PE})$:
$$ \psi_{PE} = \phi_{PE}(\mathbf{F}_{\text{emb}};\theta_{PE}) $$

\noindent \textbf{WHF Tune:} Complementary to $\psi_{PE}$, this module processes the high-frequency feature map $I_{HF}$ from the WHF module ($\S$\ref{sec:wavelet}). $I_{HF}$ can be real or complex depending on the wavelet. 
Since the features $I_{HF}$ are 2D spatial features, they are divided into patches and projected using a convolution layer to obtain  $\psi_{whf}$, matching $\psi_{PE}$'s dimensions.

\noindent \textbf{Real/Complex Adapters:} Adapters are placed within SAM's image encoder layers. For layer $i$, the adapter input is the channel concatenation $\Psi = [\psi_{PE}, \psi_{whf}] \in \mathbb{R}^{\frac{H}{16} \times \frac{W}{16} \times 2\gamma}$ (or complex). The output $\mathbf{P}^{(i)} \in \mathbb{R}^{\frac{H}{16} \times \frac{W}{16} \times C_{img}^i}$ is computed via:
$$ \mathbf{P}^{(i)} = \phi_{up}^{(i)}(\sigma(\phi_{tune}^{(i)}(\Psi))) $$
$\psi_{tune}^{(i)}$ and $\psi_{up}^{(i)}$ are MLPs (standard or complex), and the activation function ($\sigma$) is GeLU (or split GeLU for complex). $\mathbf{P}^{(i)}$ is then added with the layer's feature map $\mathbf{F}_{img}^i$:
$$ \overline{\mathbf{F}}_{img}^{(i)} = \mathbf{F}_{img}^{(i)}+ \mathbf{P}^{(i)} $$

In case of complex adapters, the real and imaginary parts of $\mathbf{P}^{(i)}$ are both added separately to $\mathbf{F}_{img}$.
The $\mathbf{P}^{(i)}$ feature is obtained for each layer separately and then added to the respective layer's feature map. This provides robustness to learn features required for adaptation.

 \noindent SAMwave benefits from: 1) \textbf{Feature Complementarity:} combining SAM's semantic features with explicit learned high-frequency details; 2) \textbf{Wavelet Flexibility:} easily integrates various wavelets, including complex ones with $\mathbb{C}$-Adapters to leverage properties like shift invariance and phase information for improved robustness on challenging tasks.

\section{Experimental Details and Analysis}
We evaluate our approach on four low-level vision tasks: camouflaged object detection, shadow detection, defocus blur detection, and polyp detection and compare the performance with current state-of-the-art methods. We also perform ablation studies to analyze the effectiveness of our approach.
\begin{table*}[h]
\caption{\small Comparison of Camouflaged Object Detection performance on three datasets against task-specific and finetuning methods. Our SAMwave approach demonstrates improved performance across all datasets compared to existing methods.(\crule[red]{.18cm}{.18cm}: best, \crule[blue]{.18cm}{.18cm}: second best)}
\label{tab: cod}
\vspace{5pt}
\centering
\resizebox{0.8\textwidth}{!}{
\begin{tabular}{l||cccc||cccc||cccc}
\toprule
\midrule
 \multirow{2}{*}{Method} & \multicolumn{4}{c||}{CHAMELEON} & \multicolumn{4}{c|}{CAMO} & \multicolumn{4}{c}{COD10K} \\
 & $\mathcal{S}_m \uparrow$    & $\mathcal{E}_\phi \uparrow$    & $\mathcal{F}^{w}_{\beta} \uparrow$    & $\mathcal{M} \downarrow$    & $\mathcal{S}_m \uparrow$   & $\mathcal{E}_\phi \uparrow$   & $\mathcal{F}^{w}_{\beta} \uparrow$   & $\mathcal{M} \downarrow$  & $\mathcal{S}_m \uparrow$    & $\mathcal{E}_\phi \uparrow$   & $\mathcal{F}^{w}_{\beta} \uparrow$   & $\mathcal{M} \downarrow$   \\ \midrule
     SINet$^{(2020)}$\cite{Fan_2020_CVPR}     & 0.869 & 0.891 & 0.740 & 0.044 & 0.751 & 0.771 & 0.606 & 0.100 & 0.771 & 0.806 & 0.551 & 0.051 \\
     RankNet$^{(2021)}$\cite{yunqiu_cod21}   & 0.846 & 0.913 & 0.767 & 0.045 & 0.712 & 0.791 & 0.583 & 0.104 & 0.767 & 0.861 & 0.611 & 0.045 \\
     JCOD$^{(2021)}$\cite{DBLP:conf/cvpr/Li0LL0D21}      & 0.870 & 0.924 & -     & 0.039 & 0.792 & 0.830 & -     & 0.082 & 0.800 & 0.872 & -     & 0.041 \\
     PFNet$^{(2021)}$\cite{DBLP:conf/cvpr/MeiJW0WF21}     & 0.882 & 0.942 & 0.810 & 0.033 & 0.782 & 0.852 & 0.695 & 0.085 & 0.800 & 0.868 & 0.660 & 0.040 \\
     FBNet$^{(2023)}$\cite{DBLP:journals/tomccap/LinT0ML23}     & 0.888 & 0.939 & 0.828 & 0.032 & 0.783 & 0.839 & 0.702 & 0.081 & 0.809 & 0.889 & 0.684 & 0.035 \\
     FSPNet$^{(2023)}$\cite{DBLP:conf/cvpr/HuangDXWCQX23}    &- & - & - & - & 0.856 & 0.899 & 0.799 & 0.050 & 0.851 & 0.895 & 0.735 & 0.026 \\
   SAM$^{(2023)}$ \cite{DBLP:conf/iccv/KirillovMRMRGXW23}& 0.727 & 0.734 & 0. 639 & 0.081 & 0.684 & 0.687 & 0.606 & 0.132 & 0.783 & 0.798 & 0.701 & 0.050\\ 
   SAM2$^{(2024)}$ \cite{ravi2024sam2} & 0.359 & 0.375 &0.115 &0.357& 0.350& 0.411& 0.079& 0.311& 0.429& 0.505& 0.115& 0.218 \\
     EVP$^{(2023)}$\cite{DBLP:conf/cvpr/LiuSPC23}       & 0.871 & 0.917 & 0.795 & 0.036 & 0.846 & 0.895 & 0.777 & 0.059 & 0.843 & 0.907 & 0.742 & 0.029 \\
     SAM-Adapter$^{(2023)}$\cite{chen2023sam} & 0.896 & 0.919 & 0.824 & 0.033 & 0.847 & 0.873 & 0.765 & 0.070 & 0.883 & 0.918 & 0.801 & 0.025 \\ 
     SAM2-Adapter$^{(2024)}$\cite{chen2024sam2adapterevaluatingadapting} & 0.915 & 0.955 & \bluetext{\textbf{0.889}} & 0.018 & 0.855 & 0.909 & 0.810 & 0.051 & 0.899 & 0.950 & 0.850 & 0.018 \\ 

     \midrule
     \multicolumn{13}{c}{\textbf{SAM as Backbone}} \\ 
     \midrule
     \multicolumn{1}{c}{\textbf{Real-Valued Adapters}} & \multicolumn{12}{c}{}\\
     \midrule
     Ours (daubechies (db))    & 0.922  & 0.947   & 0.856   & 0.023   & 0.865  & 0.903  & 0.790 & 0.057   & 0.900      & 0.936   & 0.824    & 0.021      \\
     Ours (coiflet (cf))      & {0.923}  & {0.952}   & \redtext{\textbf{0.890}}   & {0.022}   & 0.856  & 0.888  & {0.807} & 0.060     & \bluetext{\textbf{0.903}}      & {0.939}    & {0.857}     & {0.019}      \\
     Ours (haar (ha))      & 0.923  & {0.954}   & 0.866   & 0.023   & {0.868}  & {0.905}  & 0.798 & {0.056}  & 0.899      & 0.934    & 0.819     & 0.022 \\ 
     Ours (symlet (sym))      & 0.915  & 0.947   & 0.845   & 0.027   & 0.846  & 0.880  & 0.777 & 0.067     & 0.897      & 0.938    & 0.822     & 0.021      \\
     \midrule
     \multicolumn{1}{c}{\textbf{Complex-Valued Adapters}} & \multicolumn{12}{c}{}\\
     \midrule
   Ours (Symmetric-b (sy-b))   & \redtext{\textbf{0.931}}  & 0.961  & 0.883  & 0.020  & 0.856  & 0.896  & 0.797  & 0.058  & 0.902  & 0.940   & 0.843   & 0.019         \\
  Ours (Symmetric-a (sy-a))  & \bluetext{\textbf{0.928}}  & 0.957  & 0.881  & 0.019  & 0.858  & 0.893  & 0.800  & 0.057  &  0.902 & 0.939  & 0.841  & 0.019        \\
    \midrule
     \multicolumn{13}{c}{\textbf{SAM2 as Backbone}} \\ 
     \midrule
     \multicolumn{1}{c}{\textbf{Real-Valued Adapters}} & \multicolumn{12}{c}{}\\
     \midrule
    Ours (daubechies (db))      & 0.912  & 0.952  & 0.880  & {0.018}  & \bluetext{\textbf{0.870}}  & \bluetext{\textbf{0.927}}  & 0.828  & \redtext{\textbf{0.045}}  & \redtext{\textbf{0.905}}  & \bluetext{\textbf{0.956}}  & 0.861  & \redtext{\textbf{0.016}}      \\
     
     Ours (coiflet (cf))     & 0.917  & \bluetext{\textbf{0.963}}  & 0.883  & \bluetext{\textbf{0.017}}  & 0.866  & 0.923  & 0.824  & 0.048  & \redtext{\textbf{0.905}}  & \redtext{\textbf{0.957}}  & \bluetext{\textbf{0.862}}  & \redtext{\textbf{0.016}}        \\
     Ours (haar (ha))         & {{0.918}}  & {0.962}  & \bluetext{\textbf{0.889}}  & \bluetext{\textbf{0.017}}  & 0.869  & 0.926  & \bluetext{\textbf{0.829}}  
   & \bluetext{\textbf{0.047}}  & 0.900  & 0.952  & \redtext{\textbf{0.865}}  & \bluetext{\textbf{0.017}}      \\ 
     Ours (symlet (sym))          & 0.917  & 0.961  & 0.887  & \bluetext{\textbf{0.017}}  & \redtext{\textbf{0.871}}  & \redtext{\textbf{0.929}}  & \redtext{\textbf{0.831}}  & \redtext{\textbf{0.045}}  & \redtext{\textbf{0.905}}  & 0.954  & 0.859  & \redtext{\textbf{0.016}}        \\
     \midrule
     \multicolumn{1}{c}{\textbf{Complex-Valued Adapters}} & \multicolumn{12}{c}{}\\
     \midrule
     Ours (Symmetric-b (sy-b)) & 0.917 & \redtext{\textbf{0.966}} & \redtext{\textbf{0.890}} & \redtext{\textbf{0.016}} & 0.864 & 0.921 & 0.824 & 0.049 & 0.902 & 0.955 & 0.858 & \bluetext{\textbf{0.017}} \\
     Ours (Symmetric-a (sy-a)) & 0.917&0.960 &0.881 &0.019 &0.864 &0.921 &0.826 &0.048 &\bluetext{\textbf{0.903}} & 0.955 &0.856 &\redtext{\textbf{0.016}} \\
   \midrule \bottomrule
\end{tabular}
}
\end{table*}
\subsection{Datasets and Evaluation Metrics}
\noindent \textbf{Camouflaged Object Detection. } Following \cite{chen2023sam, DBLP:conf/cvpr/LiuSPC23}, we evaluate on COD10K \cite{Fan_2020_CVPR}, CAMO \cite{ltnghia-CVIU2019}, and CHAMELEON \cite{skurowski2018animal}. Metrics: S-measure ($\mathcal{S}_m$) \cite{Structure-Measure}, mean E-measure ($\mathcal{E}_\phi$), weighted F-measure ($\mathcal{F}^{w}_{\beta}$) \cite{weighted-F}, and Mean Absolute Error ($\mathcal{M}$).

\noindent \textbf{Shadow Detection.} Following \cite{chen2023sam, DBLP:conf/cvpr/LiuSPC23}, we evaluate on SBU \cite{Vicente2016LargeScaleTO} and ISTD \cite{Wang_2018_CVPR}. Metric: Balanced Error Rate (\textit{BER}).

\noindent \textbf{Defocus Blur Detection.} Following \cite{chen2023sam, DBLP:conf/cvpr/LiuSPC23}, we evaluate on CUHK \cite{DBLP:conf/cvpr/ShiXJ14} and DUT \cite{DBLP:conf/cvpr/ZhaoZ0L18}. Metrics: F-measure ($\mathcal{F}_\beta$) and Mean Absolute Error ($\mathcal{M}$).

\noindent \textbf{PolyP Detection.} Following \cite{Trinh_2024_BMVC}, we evaluate on Kvasir-SEG \cite{DBLP:conf/mmm/JhaSRHLJJ20}, CVC-ClinicDB \cite{DBLP:journals/cmig/BernalSFGRV15}, ColonDB \cite{DBLP:journals/tmi/TajbakhshGL16}, ETIS \cite{PMID:24037504}. Metrics: \textit{MDice} and \textit{mIoU}.

Additional details about the datasets, along with training objectives and additional results on forgery detection, are provided in the \textit{supplementary}.
\subsection{Implementation Details}
All experiments are implemented in PyTorch on an NVIDIA A100 GPU. We follow a similar training setting as \cite{chen2023sam, DBLP:conf/cvpr/LiuSPC23} with an image size of $1024 \times 1024$. We use pre-trained SAM and SAM2 networks as the backbone.

\subsection{Experimental Results}
\begin{figure}
\centering
\includegraphics[width=0.9\columnwidth]{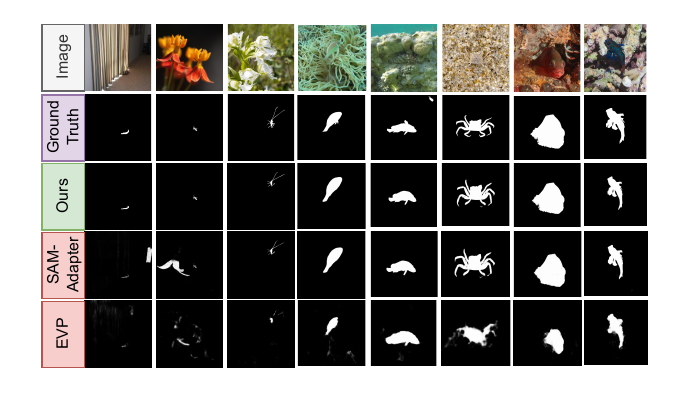}
\caption{\small Visual comparison of our proposed SAMwave with existing methods for the COD task.}
\label{fig:COD}
\end{figure}

\noindent\textbf{Camouflaged Object Detection:}
\label{sec:results_cod}
We compare our approach with existing task-specific methods \cite{DBLP:conf/cvpr/Lv0DLLBF21, DBLP:journals/tomccap/LinT0ML23, Mei_2021_CVPR} and efficient tuning methods \cite{chen2023sam, DBLP:conf/cvpr/LiuSPC23}. Tab.~\ref{tab: cod} shows that SAMwave significantly outperforms these methods across three datasets, consistently with both SAM and SAM2 backbones. To demonstrate the efficacy of our adaptive approach, we show that performance improvement over previous methods is consistent across different wavelets. Despite vanilla SAM2 performing poorly compared to SAM and other methods in Tab.~\ref{tab: cod}, its performance is significantly boosted when combined with our approach. Qualitative results are provided in Fig.~\ref{fig:COD}.



\begin{table*}[h]
\centering    
\caption{\small Comparison of state-of-the-art approaches on shadow detection.(\crule[red]{.18cm}{.18cm}: best, \crule[blue]{.18cm}{.18cm}: second best)}
\label{tab:shadow_transposed}
\vspace{5pt}
\setlength{\tabcolsep}{2pt}
\renewcommand{\arraystretch}{1.1}
\resizebox{0.9\textwidth}{!}{
\begin{tabular}{cllllll||cccccc||cccccccccccc}
\toprule \midrule
\multicolumn{7}{c||}{\multirow{4}{*}{Dataset}} & \multirow{4}{*}{\begin{tabular}[c]{@{}c@{}}DSD\\ \cite{DBLP:conf/cvpr/ZhengQCL19}\end{tabular}} & \multirow{4}{*}{\begin{tabular}[c]{@{}c@{}}FDRNet\\ \cite{DBLP:conf/iccv/Zhu0KL21}\end{tabular}} & \multirow{4}{*}{\begin{tabular}[c]{@{}c@{}}MTMT\\ \cite{DBLP:conf/cvpr/Chen0WW0H20}\end{tabular}} & \multirow{4}{*}{\begin{tabular}[c]{@{}c@{}}EVP\\ \cite{DBLP:conf/cvpr/LiuSPC23}\end{tabular}} & \multirow{4}{*}{\begin{tabular}[c]{@{}c@{}}SAM-\\ Adapter\cite{chen2023sam}\end{tabular}} & \multirow{4}{*}{\begin{tabular}[c]{@{}c@{}}SAM2-\\ Adapter\cite{chen2024sam2adapterevaluatingadapting}\end{tabular}} & \multicolumn{12}{c}{Ours}  \\ \cline{14-25} 
\multicolumn{7}{c||}{}  &   &   &  &  &   &    & \multicolumn{6}{c}{SAM as Backbone}   & \multicolumn{6}{c}{SAM2 as Backbone}   \\ \cline{14-25} 
\multicolumn{7}{c||}{}    &   &    &   &    &    &    & \multicolumn{4}{c|}{\begin{tabular}[c]{@{}c@{}}Real-valued\\ Adapters\end{tabular}}    & \multicolumn{2}{c|}{\begin{tabular}[c]{@{}c@{}}Complex-valued\\  Adapters\end{tabular}} & \multicolumn{4}{c|}{\begin{tabular}[c]{@{}c@{}}Real-valued\\ Adapters\end{tabular}}        & \multicolumn{2}{c}{\begin{tabular}[c]{@{}c@{}}Complex-valued\\ Adapters\end{tabular}} \\ \cline{14-25}
\multicolumn{7}{c||}{}      &    &    &   &   &   &   & \multicolumn{1}{c}{db} & \multicolumn{1}{c}{cf} & \multicolumn{1}{c}{ha} & \multicolumn{1}{c|}{sym} & \multicolumn{1}{c}{sy-b}  & \multicolumn{1}{c|}{sy-a}   & \multicolumn{1}{c}{db} & \multicolumn{1}{c}{cf} & \multicolumn{1}{c}{ha} & \multicolumn{1}{c|}{sym} & \multicolumn{1}{c}{sy-b}  & \multicolumn{1}{c}{sy-a}   \\ \midrule
\multicolumn{7}{c||}{ISTD (BER$\downarrow$)}   & 2.17 & 1.55 & 1.72 & 1.35 & 1.43 & 1.43 & 1.24 & 1.24 & \bluetext{\textbf{1.01}} & \multicolumn{1}{c|}{1.04} & 1.09 & \multicolumn{1}{c|}{1.15} & 1.91 & 1.24 & \redtext{\textbf{0.92}} & \multicolumn{1}{c|}{1.22} & 2.60 & 1.32 \\
\multicolumn{7}{c||}{SBU (BER$\downarrow$)}  & 3.45 & 3.04 & 3.15 & 4.31 & -    & -    & \redtext{\textbf{2.93}} & 3.01 & 3.00 & \multicolumn{1}{c|}{3.07} & 3.06 & \multicolumn{1}{c|}{3.06} & 3.04 & 5.28 & \bluetext{\textbf{2.99}} & \multicolumn{1}{c|}{3.77} & 4.43 & 4.32 \\ \midrule \bottomrule                     
\end{tabular}
}
\end{table*}

\noindent\textbf{Shadow Detection:}
We compare the performance of our approach against existing methods across two benchmark datasets. In both cases, our method yields substantial performance improvements. Although the performance improvements are relatively smaller on the ISTD dataset when using Daubechies and Coiflet wavelets compared to other wavelet variants, the improvement remains significant, demonstrating the robustness and generalizability of our adaptive approach.

\begin{table*}[h]
\centering    
\caption{\small Comparison with state-of-the-art approaches on blur detection.(\crule[red]{.18cm}{.18cm}: best, \crule[blue]{.18cm}{.18cm}: second best)}
\label{tab:blur}
\vspace{5pt}
\setlength{\tabcolsep}{2pt}
\renewcommand{\arraystretch}{1.1}
\resizebox{0.9\textwidth}{!}{
\begin{tabular}{ccllllll||cccccc||cccccccccccc}
\toprule \midrule
\multicolumn{8}{c||}{\multirow{4}{*}{Dataset}} & \multirow{4}{*}{\begin{tabular}[c]{@{}c@{}}DeFusion-\\ Net\cite{DBLP:conf/cvpr/TangZLWZ19}\end{tabular}} & \multirow{4}{*}{\begin{tabular}[c]{@{}c@{}}BTBNet\\ \cite{DBLP:conf/cvpr/ZhaoZ0L18}\end{tabular}} & \multirow{4}{*}{\begin{tabular}[c]{@{}c@{}}CENet\\ \cite{DBLP:conf/cvpr/ZhaoZLL19}\end{tabular}} & \multirow{4}{*}{\begin{tabular}[c]{@{}c@{}}DAD\\ \cite{DBLP:conf/cvpr/ZhaoSL21}\end{tabular}} & \multirow{4}{*}{\begin{tabular}[c]{@{}c@{}}EFENet\\ \cite{DBLP:journals/tip/ZhaoHHL21}\end{tabular}} & \multirow{4}{*}{\begin{tabular}[c]{@{}c@{}}EVP\\ \cite{DBLP:conf/cvpr/LiuSPC23}\end{tabular}} & \multicolumn{12}{c}{Ours}  \\ \cline{15-26} 
\multicolumn{8}{c||}{}  &   &   &  &  &   &    & \multicolumn{6}{c}{SAM as Backbone}   & \multicolumn{6}{c}{SAM2 as Backbone}   \\ \cline{15-26} 
\multicolumn{8}{c||}{}    &   &    &   &    &    &    & \multicolumn{4}{c|}{\begin{tabular}[c]{@{}c@{}}Real-valued\\ Adapters\end{tabular}}    & \multicolumn{2}{c|}{\begin{tabular}[c]{@{}c@{}}Complex-valued\\  Adapters\end{tabular}} & \multicolumn{4}{c|}{\begin{tabular}[c]{@{}c@{}}Real-valued\\ Adapters\end{tabular}}        & \multicolumn{2}{c}{\begin{tabular}[c]{@{}c@{}}Complex-valued\\ Adapters\end{tabular}} \\ \cline{15-26}
\multicolumn{8}{c||}{}      &    &    &    &    &   &   & \multicolumn{1}{c}{db} & \multicolumn{1}{c}{cf} & \multicolumn{1}{c}{ha} & \multicolumn{1}{c|}{sym} & \multicolumn{1}{c}{sy-b}  & \multicolumn{1}{c|}{sy-a}   & \multicolumn{1}{c}{db} & \multicolumn{1}{c}{cf} & \multicolumn{1}{c}{ha} & \multicolumn{1}{c|}{sym} & \multicolumn{1}{c}{sy-b}  & \multicolumn{1}{c}{sy-a}   \\ \midrule

\multirow{2}{*}{DUT}  & \multicolumn{7}{c||}{$\mathcal{F}_{\beta} \uparrow$}   & 0.823 & 0.827 & 0.817 & 0.794 & 0.854 & 0.890 & 0.888 & 0.886 & 0.880 & \multicolumn{1}{c|}{0.885} & 0.892 & \multicolumn{1}{c|}{0.893} & 0.896 & 0.893 & 0.894 & \multicolumn{1}{c|}{0.893} & \redtext{\textbf{0.907}} & \bluetext{\textbf{0.902}} \\
                   & \multicolumn{7}{c||}{$\mathcal{M} \downarrow$} & 0.118 & 0.138 & 0.135 & 0.153 & 0.094 & 0.068 & 0.068 & \bluetext{\textbf{0.067}} & 0.108 & \multicolumn{1}{c|}{0.106} & \bluetext{\textbf{0.067}} & \multicolumn{1}{c|}{0.066} & \redtext{\textbf{0.051}} & \redtext{\textbf{0.051}} & \redtext{\textbf{0.051}} & \multicolumn{1}{c|}{\redtext{\textbf{0.051}}} & \bluetext{\textbf{0.052}} & 0.056 \\ \midrule
                   
\multirow{2}{*}{CHUK}  & \multicolumn{7}{c||}{$\mathcal{F}_{\beta} \uparrow$}  & 0.818 & 0.889 & 0.906 & 0.884 & 0.914 & \redtext{\textbf{0.928}} & 0.922 & 0.916 & 0.911 & \multicolumn{1}{c|}{0.911} & 0.917 & \multicolumn{1}{c|}{0.912} & 0.923 & 0.924 & \bluetext{\textbf{0.925}} & \multicolumn{1}{c|}{0.924} & \redtext{\textbf{0.928}} & \bluetext{\textbf{0.925}} \\ 
                   & \multicolumn{7}{c||}{$\mathcal{M} \downarrow$}  & 0.117 & 0.082 & 0.059 & 0.079 & 0.053 & 0.045 & 0.040 & 0.046 & 0.070 & \multicolumn{1}{c|}{0.069} & 0.045 & \multicolumn{1}{c|}{0.046} & \redtext{\textbf{0.036}} & 0.039 & \bluetext{\textbf{0.037}} & \multicolumn{1}{c|}{0.038} & \redtext{\textbf{0.036}} & 0.039 \\ \midrule \bottomrule                     
\end{tabular}
}
\end{table*}

\noindent\textbf{Defocus Blur Detection:}
For the Blur Detection task, we primarily compare our method with EVP\cite{DBLP:conf/cvpr/LiuSPC23}, as SAM-Adapter\cite{chen2023sam} does not report results for this task. 
As shown in Tab.~\ref{tab:blur}, our approach achieves the best results when using complex-valued adapters with the SAM2 backbone. While we outperform EVP in all cases, our method demonstrates significant improvements over other existing approaches as well.

\begin{table*}[h]
\centering    
\caption{\small Comparison with state-of-the-art methods on PolyP detection. (\crule[red]{.18cm}{.18cm}: best, \crule[blue]{.18cm}{.18cm}: second best)}
\label{tab:polyp}
\vspace{5pt}
\setlength{\tabcolsep}{2pt}
\renewcommand{\arraystretch}{1.1}
\resizebox{\textwidth}{!}{
\begin{tabular}{ccllllll||cccc||cccccccccccc}
\toprule \midrule
\multicolumn{8}{c||}{\multirow{4}{*}{Dataset}} & \multirow{4}{*}{\begin{tabular}[c]{@{}c@{}}M$^2$UNet\\ \cite{DBLP:conf/eusipco/TrinhBMNPTN23}\end{tabular}} & \multirow{4}{*}{\begin{tabular}[c]{@{}c@{}}SAM\\ Adapter \cite{chen2023sam}\end{tabular}} & \multirow{4}{*}{\begin{tabular}[c]{@{}c@{}}SAM2\\ Adapter \cite{chen2024sam2adapterevaluatingadapting}\end{tabular}} & \multirow{4}{*}{\begin{tabular}[c]{@{}c@{}}SAM-EG\\ \cite{Trinh_2024_BMVC}\end{tabular}} & \multicolumn{12}{c}{Ours}  \\ \cline{13-24} 
\multicolumn{8}{c||}{}     &  &  &   &    & \multicolumn{6}{c}{SAM as Backbone}   & \multicolumn{6}{c}{SAM2 as Backbone}   \\ \cline{13-24}  
\multicolumn{8}{c||}{}       &   &    &    &    & \multicolumn{4}{c|}{\begin{tabular}[c]{@{}c@{}}Real-valued\\ Adapters\end{tabular}}    & \multicolumn{2}{c|}{\begin{tabular}[c]{@{}c@{}}Complex-valued\\  Adapters\end{tabular}} & \multicolumn{4}{c|}{\begin{tabular}[c]{@{}c@{}}Real-valued\\ Adapters\end{tabular}}        & \multicolumn{2}{c}{\begin{tabular}[c]{@{}c@{}}Complex-valued\\ Adapters\end{tabular}} \\ \cline{13-24} 
\multicolumn{8}{c||}{}           &    &    &   &   & \multicolumn{1}{c}{db} & \multicolumn{1}{c}{cf} & \multicolumn{1}{c}{ha} & \multicolumn{1}{c|}{sym} & \multicolumn{1}{c}{sy-b}  & \multicolumn{1}{c|}{sy-a}   & \multicolumn{1}{c}{db} & \multicolumn{1}{c}{cf} & \multicolumn{1}{c}{ha} & \multicolumn{1}{c|}{sym} & \multicolumn{1}{c}{sy-b}  & \multicolumn{1}{c}{sy-a}   \\ \midrule

\multirow{2}{*}{ClinicDB}  & \multicolumn{7}{c||}{mDice$\uparrow$}   & \bluetext{\textbf{0.901}} & - & - & \redtext{\textbf{0.931}} & 0.855 & 0.652 & 0.866 & \multicolumn{1}{c|}{0.854} & 0.832 & \multicolumn{1}{c|}{0.840} & 0.887 & 0.881 & 0.885 & \multicolumn{1}{c|}{0.874} & 0.862 & 0.846 \\

                   & \multicolumn{7}{c||}{mIoU$\uparrow$}   & \bluetext{\textbf{0.853}} & - & - & \redtext{\textbf{0.879}} & 0.789 & 0.449 & 0.794 & \multicolumn{1}{c|}{0.786} & 0.761 & \multicolumn{1}{c|}{0.772} & 0.826 & 0.821 & 0.827 & \multicolumn{1}{c|}{0.816} & 0.800 & 0.784 \\ \midrule

\multirow{2}{*}{ColonDB}  & \multicolumn{7}{c||}{mDice$\uparrow$}  & 0.767 & - & - & 0.774 & 0.743 & 0.635 & 0.724 & \multicolumn{1}{c|}{0.737} & 0.726 & \multicolumn{1}{c|}{0.748} & 0.764 & 0.769 & 0.769 & \multicolumn{1}{c|}{0.778} & \bluetext{\textbf{0.781}} & \redtext{\textbf{0.782}} \\

                   & \multicolumn{7}{c||}{mIoU$\uparrow$}  & 0.684 & - & - & 0.689 & 0.668 & 0.386 & 0.647 & \multicolumn{1}{c|}{0.661} & 0.652 & \multicolumn{1}{c|}{0.671} & 0.696 & 0.703 & 0.697 & \multicolumn{1}{c|}{0.704} & \redtext{\textbf{0.710}} & \bluetext{\textbf{0.707}} \\ \midrule

\multirow{2}{*}{Kvasir}  & \multicolumn{7}{c||}{mDice$\uparrow$} & 0.907 & 0.850 & 0.873 & \bluetext{\textbf{0.915}} & 0.889 & 0.812 & 0.901 & \multicolumn{1}{c|}{0.900} & 0.901 & \multicolumn{1}{c|}{0.902} & \redtext{\textbf{0.917}} & 0.913 & 0.911 & \multicolumn{1}{c|}{0.908} & 0.899 & 0.902 \\
                   & \multicolumn{7}{c||}{mIoU$\uparrow$}  & 0.855 & 0.776 & 0.806 & \bluetext{\textbf{0.862}} & 0.818 & 0.618 & 0.827 & \multicolumn{1}{c|}{0.828} & 0.834 & \multicolumn{1}{c|}{0.838} & \redtext{\textbf{0.867}} & 0.861 & 0.859 & \multicolumn{1}{c|}{0.860} & 0.844 & 0.847 \\ \midrule
                   
\multirow{2}{*}{ETIS}  & \multicolumn{7}{c||}{mDice$\uparrow$}  & 0.670 & - & - & 0.757 & 0.688 & 0.445 & 0.685 & \multicolumn{1}{c|}{0.693} & 0.680 & \multicolumn{1}{c|}{0.732} & 0.780 & 0.749 & 0.784 & \multicolumn{1}{c|}{\redtext{\textbf{0.802}}} & \bluetext{\textbf{0.797}} & 0.750 \\

                   & \multicolumn{7}{c||}{mIoU$\uparrow$}  & 0.595 & - & - & 0.681 & 0.610 & 0.223 & 0.603 & \multicolumn{1}{c|}{0.622} & 0.611 & \multicolumn{1}{c|}{0.654} & 0.709 & 0.683 & 0.720 & \multicolumn{1}{c|}{\redtext{\textbf{0.737}}} & \redtext{\textbf{0.730}} & 0.676 \\ \midrule \bottomrule                     
\end{tabular}
}
\end{table*}

\noindent\textbf{PolyP Detection:}
Tab.~\ref{tab:polyp} presents quantitative results for the PolyP detection task. Compared to existing approaches\cite{Trinh_2024_BMVC, chen2024sam2adapterevaluatingadapting, chen2023sam,DBLP:conf/eusipco/TrinhBMNPTN23}, our method achieves significant improvements across nearly all datasets. Notably, we observe further gains when using the SAM2 backbone, especially in conjunction with complex-valued adapters. While our performance is comparable to that of~\cite{Trinh_2024_BMVC}, it is important to emphasize that--unlike their task-specific model--our adaptive approach is designed to generalize across multiple low-level vision tasks.

\noindent \textbf{$\mathbb{R}$-valued \textit{vs} $\mathbb{C}$-valued Adapters:} Tab.~\ref{tab: cod},~\ref{tab:shadow_transposed},~\ref{tab:blur}, and~\ref{tab:polyp} show that, when using the SAM backbone, complex-valued adapters perform comparably to—or in some cases better than—their real-valued adapters. Furthermore, when switching to the SAM2 backbone, complex-valued adapters consistently yield improved results, highlighting the enhanced compatibility and representational capacity of SAM2 for this task.


\colourcheck{green}
\begin{table}[ht]
\vspace{-5pt}
\caption{\small The results of our ablation study highlight the importance of adapters for efficient finetuning; it also highlights the importance of each low-frequency subband (\textit{LL}) and the rest of the high-frequency subbands (\textit{HL}, \textit{LH}, \textit{HH}) in the image in the wavelet domain. The ablation experiment is done for camouflage object detection, and SAM is taken as the backbone.}
\label{tab:ablation}
\centering
\vspace{5pt}
\resizebox{0.9\textwidth}{!}{
\begin{tabular}{cccc||cccc||cccc||cccc}
\toprule
\midrule
\multicolumn{4}{c||}{ Method}& \multicolumn{4}{c||}{CHAMELEON} & \multicolumn{4}{c||}{CAMO} & \multicolumn{4}{c}{COD10K} \\
                    & &  &  & $\mathcal{S}_m \uparrow$    & $\mathcal{E}_\phi \uparrow$    & $\mathcal{F}^{w}_{\beta} \uparrow$    & $\mathcal{M} \downarrow$    & $\mathcal{S}_m \uparrow$   & $\mathcal{E}_\phi \uparrow$   & $\mathcal{F}^{w}_{\beta} \uparrow$   & $\mathcal{M} \downarrow$  & $\mathcal{S}_m \uparrow$    & $\mathcal{E}_\phi \uparrow$   & $\mathcal{F}^{w}_{\beta} \uparrow$   & $\mathcal{M} \downarrow$   \\ \midrule
     \multicolumn{4}{c||}{SAM Finetuned} & 0.796 & 0.802 & 0.676 & 0.062 & 0.750 & 0.756 & 0.639 & 0.105 & 0.789 & 0.817& 0.596 & 0.049\\
      \midrule \midrule
      \textit{LL} & \textit{HL} & \textit{LH} & \textit{HH} & \multicolumn{12}{c}{Evaluating each component} \\ 
      \midrule
     \textcolor{black}{\ding{52}} & \cross & \textcolor{red}{\ding{55}} & \cross   & 0.907 & 0.937 & 0.842 & 0.029 & \textbf{0.856} & 0.884 & 0.785 & 0.062 & 0.891 & 0.929 & 0.817 & 0.024 \\
     \cross & \textcolor{black}{\ding{52}}& \textcolor{red}{\ding{55}} & \cross   & 0.895 & 0.927 & 0.834 & 0.030 & 0.851 & 0.882 & 0.781 & 0.063 & 0.887 & 0.924 & 0.816 & 0.024 \\
    \cross & \cross & \textcolor{black}{\ding{52}}& \cross   & 0.896 & 0.915 & 0.838 & 0.030 & 0.854 & 0.887 & 0.790 & 0.063 & 0.891 & 0.928 & 0.838 & 0.023 \\
     \cross & \cross & \cross & \textcolor{black}{\ding{52}}& 0.908 & 0.935 & 0.848     & 0.028 & 0.853 & 0.883 & 0.783     & 0.064 & 0.893 & 0.928 & 0.818     & 0.023 \\
     \midrule
     \cross & \textcolor{black}{\ding{52}}& \textcolor{black}{\ding{52}}& \textcolor{black}{\ding{52}}&  \textbf{0.923}  & \textbf{0.952}   & \textbf{0.890}   & \textbf{0.022}   & \textbf{0.856}  & \textbf{0.888}  & \textbf{0.807} & \textbf{0.060}     & \textbf{0.903}      & \textbf{0.939}    & \textbf{0.857}     & \textbf{0.019}       \\

    \midrule
     \bottomrule
\end{tabular}
}
\vspace{-25pt}
\end{table}

\subsection{Ablation Study}
We conduct three ablation studies to highlight our contribution. First, we remove the WHF module and adapters from the SAM backbone while keeping everything else the same, and then we provide each of the subbands of wavelets transform. Lastly, we replace the wavelets used in our approach with other wavelets and observe the result.

\textbf{Effect of adapters:} We remove high-frequency features and patch embeddings from the backbone and train the SAM model. Tab.~\ref{tab:ablation} shows the result for the same. Although this helped improve performance over SAM, it lags significantly.

\textbf{Effect of each subband:} In $\S$\ref{sec:wavelet}, we claim that when combining all three high-frequency subbands (\textit{HL}, \textit{LH}, \textit{HH}), there is a significant improvement in the performance. Hence, we perform an ablation study by considering each subband individually and observing their significance to assert our claim. As expected, when provided with low-frequency features (\textit{LL}), we see performance degradation for all three datasets. Similarly, when we provide each high-frequency subband, we see a similar pattern in performance loss. However, we only see the best performance when all three are combined. 

\textbf{Effect of each wavelet:} To ensure that the results obtained are not due to any specific property of a single wavelet, we use four different wavelets and show their results. The comparative result are shown in Tab.~\ref{tab: cod}, \ref{tab:shadow_transposed}, \ref{tab:blur}, \ref{tab:polyp}. The observation is similar across all four tasks; we see more performance gain for some wavelets, but overall, using any wavelet provides better finer detail information than previous methods. 
\section{Conclusion}
This work presents a new efficient fine-tuning method for large vision models, particularly SAM and SAM2, for solving four different low-level vision tasks. Additionally, we show that a complex-valued adapter can be used to fine-tune large vision models efficiently.
We first show that the existing method of providing high-frequency information to the network is not optimal and needs improvement. We propose a WHF module, which converts the input image to a wavelet domain and combines the high-frequency subbands. When provided to the networks, this new feature map shows significant improvement over existing methods.

\bibliography{main}
\newpage
\appendix
\section{Preliminaries}
\subsection{Wavelet Transform}
Wavelet Transform is a classical image processing technique widely used in image compression to separate the low-frequency approximation and the high-frequency details from the original image. Although the original idea of wavelets was proposed for continuous signals, discrete signals like images can also be transformed using discrete wavelet transform (DWT) and inverse wavelet transform (IWT) to regain the original discrete image.
To understand DWT, let us start with an input image ${I}\in \mathbb{R}^{H\times W\times 3}$, a low pass filter $L$, and a high pass filter $H$, where:

\begin{equation*}
L = \frac{1}{\sqrt{2}}\begin{bmatrix}
    1 & 1\\
\end{bmatrix}   \textrm{ \& }  H = \frac{1}{\sqrt{2}}\begin{bmatrix}
    1 & -1\\
\end{bmatrix}
\end{equation*}

Now, using these two filters, a 2D kernel can be constructed by applying each one row-wise and column-wise. It will result in four kernels of stride 2: $LL^T$, $LH^T$, $HL^T$, $HH^T$. 
Using the kernels, we decompose the input ${I}$ into four subbands of size $H/2 \times W/2$, i.e.,  ${I}_{ll}$, ${I}_{lh}$, ${I}_{hl}$, ${I}_{hh}$.
Note that the kernels are pairwise orthogonal; a $4\times 4$ invertible matrix can be formed to reconstruct the original image accurately from its low-resolution frequency subbands by IWT. 

For complex wavelet transform, we replace real-valued wavelets with complex-valued wavelets. Generally, it is also implemented using a dual-tree form focusing on the real and imaginary components of the complex-valued output.
\section{Additional Method details}

\subsection{Training Objectives}
\label{sec:lossfun}
Following \cite{DBLP:conf/cvpr/LiuSPC23, chen2023sam}, we optimize the tasks in SAMWave by minimizing the loss between the predicted binary mask ($\hat{m}$) and ground truth ($m$). Specifically, we use binary cross-entropy loss (${L}_{bce}$)\cite{DBLP:journals/anor/BoerKMR05} for defocus blur detection and forgery detection, balanced binary cross-entropy loss (${L}_{bbce}$) for shadow detection and summation of binary cross-entropy and intersection-over-union loss (${L}_{bbce} + {L}_{iou}$) for camouflaged object detection. The loss functions are formulated below as follows:

\begin{equation}
    \mathcal{L}_{bce}(\hat{m}, m) = -\big(m*log(\hat{m}) 
                      + (1 - m)*log(1 - \hat{m})\big)
\end{equation}
\begin{equation}
\begin{split}
    \mathcal{L}_{bbce}(\hat{m}, m) &= -\zeta*\big(m*log(\hat{m}) \\ 
                      &+ (1 - m)*log(1 - \hat{m})\big)
\end{split}
\end{equation}
\begin{equation}
    \zeta = \frac{\sum m + \epsilon}{\sum m + \sum (1 - m) + \epsilon}
\end{equation}
\begin{equation}
\begin{split}
    \mathcal{L}_{iou}(\hat{m}, m) &= 1 - \frac{\sum(\hat{m}*m)}{\sum(\hat{m}+m) - \sum(\hat{m}*m)}
\end{split}
\end{equation}

\section{Additional Experimental Results}
\subsection{Datasets}
\label{sec:dataset}
The train and test split for all tasks are shown in Tab.~\ref{tab:data}.
\begin{table}[ht]
\caption{Overview of the datasets used for training and evaluation of our proposed method. In the following table if the training images are ``-", then the dataset is only used for testing.}
\vspace{5pt}
\label{tab:data}
\resizebox{\columnwidth}{!}{
\begin{tabular}{l|c|cc}
\toprule
Tasks                                         & Datasets  & Training Images & Testing Images \\ \midrule
\multirow{2}{*}{Forgery Detection}            & CASIA \cite{6625374}     & 5213            & 921            \\
                                              & IMD20 \cite{9096940}    & -               & 2010           \\ \midrule
\multirow{2}{*}{Shadow Detection}             & ISTD \cite{Wang_2018_CVPR}     & 1330            & 540            \\
                                              & SBU \cite{Vicente2016LargeScaleTO}      & 4089            & 638            \\ \midrule
\multirow{2}{*}{Defocus Blur Detection}       & CUHK\cite{DBLP:conf/cvpr/ShiXJ14}      & 604             & 100            \\
                                              & DUT\cite{DBLP:conf/cvpr/ZhaoZ0L18}       & -               & 500            \\ \midrule
\multirow{3}{*}{Camouflaged Object Detection} & COD10K \cite{Fan_2020_CVPR}   & 3040            & 2026           \\
                                              & CAMO \cite{ltnghia-CVIU2019}     & 1000            & 250            \\
                                              & CHAMELEON\cite{skurowski2018animal} & -               & 76             \\ \midrule
\multirow{4}{*}{PolyP Detection}              & CVC-ClinicDB \cite{DBLP:journals/cmig/BernalSFGRV15}   & 550            & 62           \\
                                              & ColonDB \cite{DBLP:journals/tmi/TajbakhshGL16}     & -            & 380            \\
                                              & Kvasir-SEG \cite{DBLP:conf/mmm/JhaSRHLJJ20}     & 900            & 100            \\
                                              & ETIS\cite{PMID:24037504} & -               & 196             \\ \bottomrule
\end{tabular}
}
\label{tab:data}
\end{table}
\subsection{Forgery Detection}
Following \cite{chen2023sam, DBLP:conf/cvpr/LiuSPC23}, we evaluate on CASIA \cite{6625374} and IMD20 \cite{9096940}. Metrics: pixel-level AUC and $\mathcal{F}_1$ score.
\begin{table*}[h]
\centering
\caption{\small Comparison of state-of-the-art approaches on forgery detection.(\crule[red]{.18cm}{.18cm}: best, \crule[blue]{.18cm}{.18cm}: second best)}
\label{tab:forgery}
\vspace{5pt}
        \setlength{\tabcolsep}{7pt}
\renewcommand{\arraystretch}{1}
        \resizebox{.5\linewidth}{!}{
        
        \begin{tabular}{lcc}
        \toprule
        \midrule
        \multirow{2}{*}{Method} & \multicolumn{2}{c}{IMD20} \\
                                & $\mathcal{F}1 \uparrow$    & AUC $\uparrow$   \\ 
        \midrule
         ManTra$^{(2019)}$\cite{DBLP:conf/cvpr/0001AN19}              & -      & 0.748     \\
         SPAN$^{(2020)}$\cite{DBLP:conf/eccv/HuZJCYN20}                & -      & 0.750         \\
         PSCCNet$^{(2022)}$\cite{liu2022pscc}             & -      & 0.806         \\
         TransForensics$^{(2021)}$\cite{DBLP:conf/iccv/HaoZYXP21}      & -      & 0.848         \\
         ObjectFormer$^{(2022)}$\cite{DBLP:conf/cvpr/WangWCHSLJ22}        & -      & 0.821         \\
         EVP$^{(2023)}$\cite{DBLP:conf/cvpr/LiuSPC23}                 & 0.443  & 0.807         \\ 
    \midrule
    \multicolumn{3}{c}{\textbf{Real-valued Adapters}} \\ 
    \midrule
          \multicolumn{3}{c}{SAM as backbone}\\
      \midrule
         Ours (daubechies)               &   {0.539}     & {0.874}     \\ 
         Ours (coiflet)               & \redtext{\textbf{0.626}}       & \bluetext{\textbf{0.902}}     \\
         Ours (haar)               &  \bluetext{\textbf{0.614}}      & \redtext{\textbf{0.903}}      \\
         Ours (symlet)               & 0.607       & 0.900     \\
    \midrule
          \multicolumn{3}{c}{SAM2 as backbone}\\
      \midrule
         Ours (daubechies)               &   {0.132}     & {0.500}     \\ 
         Ours (coiflet)               & {0.416}       & {0.658}     \\
         Ours (haar)               &  {0.131}      & {0.903}      \\
         Ours (symlet)               & 0.607       & 0.900     \\
          \midrule
         \bottomrule
        \end{tabular}
    }
\end{table*}
We compare our approach with existing task-specific methods \cite{DBLP:conf/cvpr/0001AN19, DBLP:conf/eccv/HuZJCYN20, liu2022pscc, DBLP:conf/iccv/HaoZYXP21, DBLP:conf/cvpr/WangWCHSLJ22} and efficient tuning methods \cite{DBLP:conf/cvpr/LiuSPC23}. As shown in Tab.~\ref{tab:forgery}, our adaptive approach significantly outperforms prior approaches across the board. Notably, using the SAM backbone in conjunction with \textit{Coiflet} and \textit{Haar} wavelets, our adaptive framework achieves particularly strong performance, demonstrating its effectiveness in high-frequency, detail-sensitive tasks such as image forgery detection.

\end{document}